\documentclass[conference, a4paper, 10pt]{IEEEtran}
\usepackage{graphicx}
\usepackage{amsmath}
\usepackage{cite}
\usepackage{url}
\usepackage{booktabs}
\usepackage{multirow}
\usepackage{color}
\usepackage{xspace}
\usepackage{hyperref}
\usepackage{amssymb}
\usepackage{microtype}  
\usepackage[caption=false,font=footnotesize]{subfig}

\title{OmicsCL: Unsupervised Contrastive Learning for Cancer Subtype Discovery and Survival Stratification}

\author{
    \IEEEauthorblockN{\large Atahan Karagöz}
    \IEEEauthorblockA{
        \textit{Department of Computer Science} \\
        \textit{University of Basel} \\
        Basel, Switzerland \\
        atahan.karagoez@stud.unibas.ch
    }
}

\begin{document}

\maketitle

\begin{abstract}
    Unsupervised learning of disease subtypes from multi-omics data presents a significant opportunity for advancing personalized medicine. We introduce \textit{OmicsCL}, a modular contrastive learning framework that jointly embeds heterogeneous omics modalities—such as gene expression, DNA methylation, and miRNA expression—into a unified latent space. Our method incorporates a survival-aware contrastive loss that encourages the model to learn representations aligned with survival-related patterns, without relying on labeled outcomes. Evaluated on the TCGA BRCA dataset, \textit{OmicsCL} uncovers clinically meaningful clusters and achieves strong unsupervised concordance with patient survival. The framework demonstrates robustness across hyperparameter configurations and can be tuned to prioritize either subtype coherence or survival stratification. Ablation studies confirm that integrating survival-aware loss significantly enhances the predictive power of learned embeddings. These results highlight the promise of contrastive objectives for biological insight discovery in high-dimensional, heterogeneous omics data.
\end{abstract}
\vspace{1em}

\section{Introduction}

Cancer is a heterogeneous disease that manifests through complex molecular alterations across different biological levels, including genomics, epigenomics, and transcriptomics. The advent of high-throughput sequencing technologies has enabled researchers to collect multi-omics datasets, which together provide a comprehensive molecular view of individual tumors. Integrating these heterogeneous data types is critical to uncovering latent subtypes and stratifying patients based on prognosis. However, effective multi-omics integration remains a challenging task due to the high dimensionality, noise, and inconsistency across omics modalities.

Traditional approaches to subtype discovery have relied on supervised learning using predefined cancer subtype labels, or unsupervised clustering techniques with limited biological interpretability. More recently, deep learning has shown promise in learning compact representations of omics data, yet many models are either supervised or require complex architectures and large amounts of annotated data. Additionally, models that do not explicitly incorporate survival outcomes may fail to capture clinically relevant distinctions between subtypes.

To address these challenges, we propose \textbf{OmicsCL}, an unsupervised contrastive learning framework designed to learn joint embeddings from multi-omics data with no subtype labels. \textit{OmicsCL} integrates contrastive objectives across omics-specific encoders while incorporating a novel survival-aware contrastive loss that encourages embeddings of patients with similar survival outcomes to be closer in latent space. This design enables the model to learn biologically meaningful and survival-informative representations in a purely unsupervised manner.

\section{Related Work}

The integration of multi-omics data for cancer subtype discovery has been extensively studied over the past decade. Traditional methods such as Similarity Network Fusion (SNF) \cite{wang2014similarity} and iCluster \cite{shen2009integrative} combine heterogeneous data sources to build unified representations, often followed by unsupervised clustering. While effective, these approaches rely on predefined similarity metrics and do not directly optimize for downstream survival relevance.

With the rise of deep learning, autoencoder-based methods have become popular for learning low-dimensional representations from high-dimensional omics data. Models such as MOFA \cite{argelaguet2018multi} and DCCA \cite{andrew2013deep} leverage probabilistic or canonical correlation-based frameworks to jointly embed multiple modalities. However, these approaches typically assume paired data distributions and are not inherently designed for survival-aware representation learning.

Contrastive learning has recently emerged as a powerful unsupervised method for representation learning in various domains, including computer vision and bioinformatics. Methods such as scCL \cite{du2023scccl} and CONAN \cite{ke2021conan} apply contrastive objectives to single-cell or multi-omics settings. Most of these methods focus on learning modality-invariant features or maximizing agreement across data views, but they often neglect the temporal aspect of clinical outcomes like patient survival.

Survival analysis in deep learning has traditionally been addressed through supervised models such as DeepSurv \cite{katzman2018deepsurv} or DeepHit \cite{lee2018deephit}, which require labeled event times and typically predict survival functions directly. While these models have achieved strong performance, they require extensive labeled data and are not naturally suited for unsupervised stratification tasks. Recent large-scale benchmarks \cite{herrmann2021large} have emphasized the limitations of supervised survival models under multi-omics settings, reinforcing the need for more flexible, unsupervised alternatives.

Our work bridges the gap between unsupervised representation learning and survival analysis. Unlike previous models, \textit{OmicsCL} introduces a survival-aware contrastive loss that encodes temporal outcome information into the embedding space without relying on explicit survival supervision or pre-defined subtype labels. This allows for discovery of clinically meaningful cancer subtypes in an entirely label-free setting, while still preserving discriminative features for patient stratification.

\section{Methodology}

\subsection{Problem Definition}

Let $\mathcal{D} = \{(\mathbf{x}^{(g)}_i, \mathbf{x}^{(m)}_i, \mathbf{x}^{(r)}_i, t_i, e_i)\}_{i=1}^N$, represent a multi-omics dataset consisting of $N$ patients, where $\mathbf{x}^{(g)}_i$, $\mathbf{x}^{(m)}_i$, and $\mathbf{x}^{(r)}_i$ correspond to gene expression, DNA methylation, and miRNA profiles, respectively. Each patient also has an associated survival time $t_i \in \mathbb{R}^{+}$ and an event indicator $e_i \in \{0, 1\}$, with $1$ denoting death and $0$ indicating censoring. Our objective is to learn compact embeddings for each omics modality and a joint representation that enables meaningful clustering of patients into subtypes predictive of survival outcomes, without relying on subtype labels during training.

\subsection{Model Architecture}

As a modular contrastive learning framework, \textit{OmicsCL} learns view-specific and joint representations across omics modalities. For each modality, we use a dedicated encoder $f_\theta^{(v)}$ parameterized by neural networks with shared structure but independent weights. Each encoder consists of a multi-layer perceptron with batch normalization and ReLU activations, followed by a projection head that maps features to a latent space $\mathbb{R}^d$, where $d$ is the embedding dimension. These projections are $\ell_2$-normalized to lie on the unit hypersphere.

\subsection{Contrastive Objective Across Omics}

Given a minibatch of samples, we construct positive pairs $(z_i^{(v)}, z_i^{(w)})$ for each $i$ from different modalities $v \neq w$, and negative pairs $(z_i^{(v)}, z_j^{(w)})$ for $j \neq i$. We adopt the normalized temperature-scaled cross-entropy loss (NT-Xent) \cite{chen2020simple}:

\begin{equation}
\mathcal{L}_{\text{NT-Xent}} = -\sum_{i=1}^{N} \log \frac{\exp(\text{sim}(z_i^{(v)}, z_i^{(w)}) / \tau)}{\sum_{j=1}^{N} \mathbb{1}_{[j \neq i]} \exp(\text{sim}(z_i^{(v)}, z_j^{(w)}) / \tau)},
\end{equation}

where $\text{sim}(a, b)$ denotes cosine similarity, and $\tau$ is a temperature parameter. This objective encourages agreement between representations from different omics views of the same patient.

\subsection{Survival-Aware Contrastive Loss}

To encode temporal risk structure into the embedding space, we introduce a novel unsupervised \textit{survival contrastive loss}. It penalizes representations of patients with dissimilar survival times (when both are deceased) and encourages closeness between embeddings with similar outcomes. Let $d_{ij}$ denote the Euclidean distance between embeddings $z_i$ and $z_j$, and $\Delta t_{ij} = |t_i - t_j|$ their survival time difference. The loss is defined as:

\begin{align}
    \mathcal{L}_{\text{surv}} =\,
    & \lambda_{\text{pull}} \cdot \mathbb{E}_{i,j} \left[ \mathbb{1}_{[e_i = e_j = 1]} \cdot \mathbb{1}_{[\Delta t_{ij} < \delta]} \cdot d_{ij}^2 \right] \nonumber \\
    & \, + \lambda_{\text{push}} \cdot \mathbb{E}_{i,j} \left[ \mathbb{1}_{[\Delta t_{ij} \geq \delta]} \cdot \max(0, \delta - d_{ij})^2 \right],
\end{align}    

where $\delta$ is a tunable time margin, and $\lambda_{\text{pull}}, \lambda_{\text{push}}$ are weighting coefficients. Notably, this formulation does not rely on supervised risk labels and operates in an entirely unsupervised regime, allowing it to generalize across cancer types and data splits.

\subsection{Joint Training}

The final training objective is a weighted sum of contrastive loss and the survival-aware regularization term:

\begin{equation}
\mathcal{L}_{\text{total}} = \mathcal{L}_{\text{NT-Xent}} + \alpha \cdot \mathcal{L}_{\text{surv}},
\end{equation}

where $\alpha$ controls the trade-off between modality alignment and survival stratification. During training, we utilize a cyclical learning rate scheduler and early stopping based on validation concordance index (C-index), which evaluates how well the learned embeddings capture survival risks.

\subsection{Clustering and Evaluation}

After training, we concatenate the learned embeddings across omics modalities to form a unified patient representation. These representations are clustered using KMeans. Evaluation is performed using clustering and survival metrics, which we detail in Section~\ref{sec:evaluation-metrics}.

\section{Experiments}

\subsection{Dataset}

We evaluated our proposed framework on the TCGA-BRCA dataset from the Multi-Omics Cancer Benchmark \cite{leng2022benchmark}. The dataset included three primary omics views: gene expression (RNA-seq), DNA methylation (450k array), and miRNA expression. Survival information was provided for each patient, including time to event or censoring and an event indicator. Additionally, PAM50 subtype annotations were available for benchmarking clustering performance.

\subsection{Data Preprocessing}

We harmonized sample identifiers across omics sources and removed patients with missing survival time or event status. Survival times were extracted from clinical fields such as ``overall\_survival,'' with corresponding binary death indicators derived from ``status''; both are consistently encoded. Subtype labels with missing values were imputed as ``Unknown'' and excluded from supervised evaluation metrics. All preprocessing scripts were released as part of our pipeline for reproducibility.

After preprocessing, the dataset comprised a total of 612 patients with all three omics modalities and survival labels available. We applied z-score normalization to each omics view and split the dataset into 60\% training, 20\% validation, and 20\% test sets using a fixed random seed for reproducibility.

\subsection{Implementation Details}

Each omics encoder was a two-layer MLP with hidden dimension $128$ and projection dimension $64$. All embeddings were $\ell_2$-normalized. The models were trained using the Adam optimizer with a weight decay of $1 \times 10^{-6}$ and a cyclical learning rate policy ranging from $1 \times 10^{-5}$ to $1 \times 10^{-3}$. The NT-Xent loss temperature $\tau$ was set to 0.1. The survival contrastive loss used a margin of 1.0, and the weighting coefficient $\alpha$ was set to 10.0 based on grid search.

\begin{figure}[ht]
    \centering
    \includegraphics[width=0.9\linewidth]{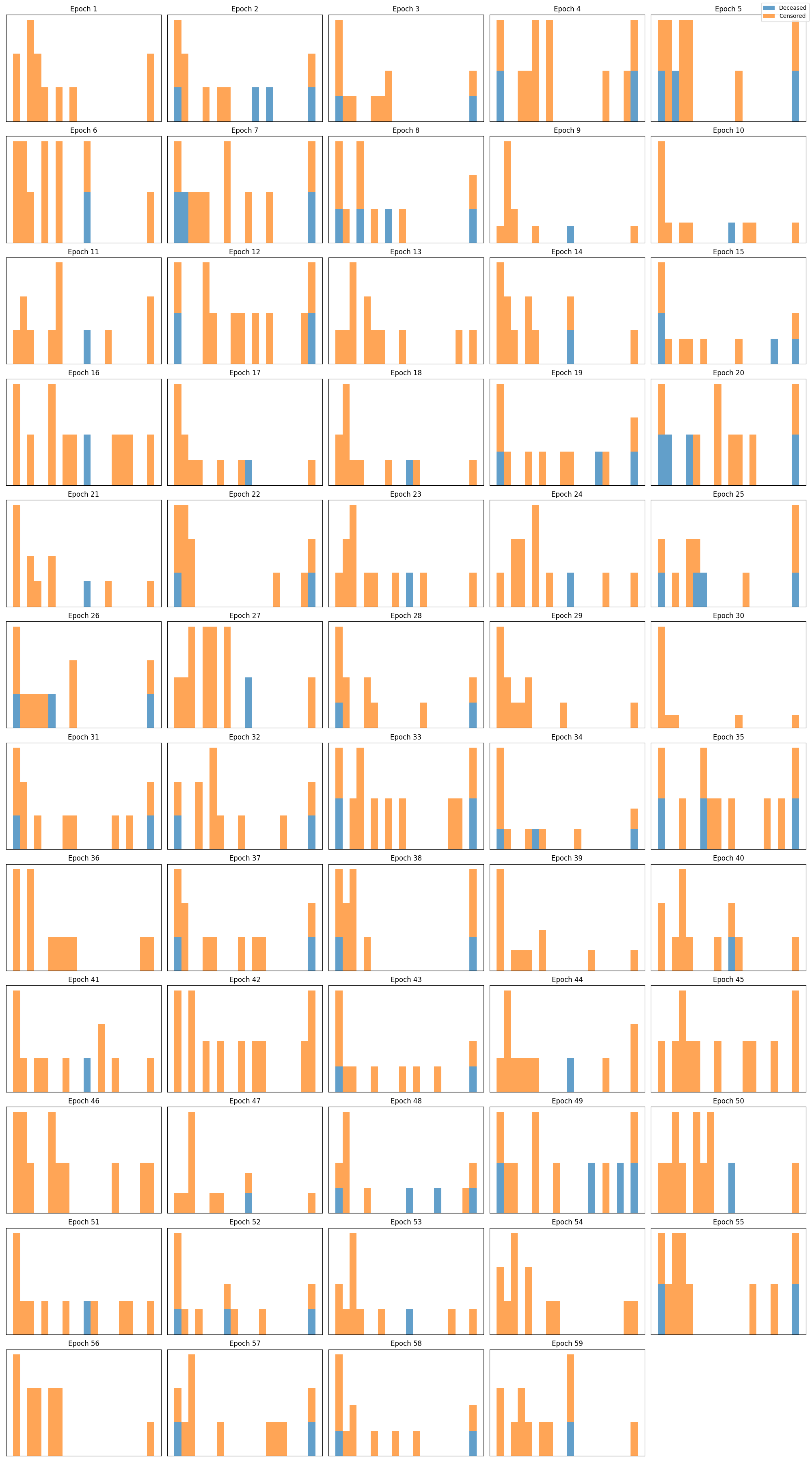}
    \caption{Combined survival histograms over training epochs. This visualization captures the evolution of censored and deceased event distributions throughout training, highlighting the temporal dynamics our model encodes into the embedding space.}
    \label{fig:survival_histograms}
\end{figure}

Training proceeded for a maximum of 1000 epochs with early stopping based on validation concordance index, using a patience of 20 epochs. The temporal dynamics captured during training are visualized in Figure~\ref{fig:survival_histograms}. Each training run was seeded for reproducibility. All experiments were run on a machine with an Apple M1 Max CPU and 64GB RAM.

\subsection{Evaluation Metrics}
\label{sec:evaluation-metrics}

To assess the quality of learned representations, we evaluate both clustering coherence and survival relevance of the predicted clusters.

\paragraph{Clustering Metrics} To evaluate agreement between predicted clusters and known PAM50 subtypes, we report several unsupervised clustering metrics. \textit{Silhouette Score} measures the cohesion and separation of samples within clusters. \textit{Purity} reflects the proportion of correctly assigned samples based on majority voting in each cluster. \textit{Adjusted Rand Index (ARI)} quantifies similarity between predicted and true labels, adjusted for chance. \textit{Normalized Mutual Information (NMI)} measures the shared information between cluster assignments and ground truth subtypes.

\paragraph{Survival Metrics} To assess the ability of clusters to stratify patient survival, we use survival-specific metrics. The \textit{Concordance Index (C-index)} evaluates the agreement between predicted risk scores and actual survival times. The \textit{log-rank test} provides a statistical measure of survival separation across clusters. Finally, \textit{Kaplan--Meier curves} visualize survival probabilities over time for each predicted cluster.

\subsection{Baselines}

As our primary goal is to remain unsupervised, we compared our method’s performance to a Cox Proportional Hazards (CoxPH) model trained on the same features. However, it should be noted that CoxPH directly optimizes supervised survival prediction, whereas \textit{OmicsCL} infers survival-relevant embeddings without access to labels.

We also compared against ablated versions of \textit{OmicsCL} trained without the survival-aware contrastive regularization, showing the impact of our design choices on downstream survival analysis.

\section{Results}

\subsection{Survival Stratification Performance}

\textit{OmicsCL} achieved strong performance in survival prediction, evidenced by a concordance index (C-index) of \textbf{0.7512} on the test set. This result demonstrated that the unsupervised embeddings learned by the model effectively capture risk-related structure in the patient population, despite the absence of label supervision during training.

\begin{figure}[!ht]
    \centering
    \includegraphics[width=0.9\linewidth]{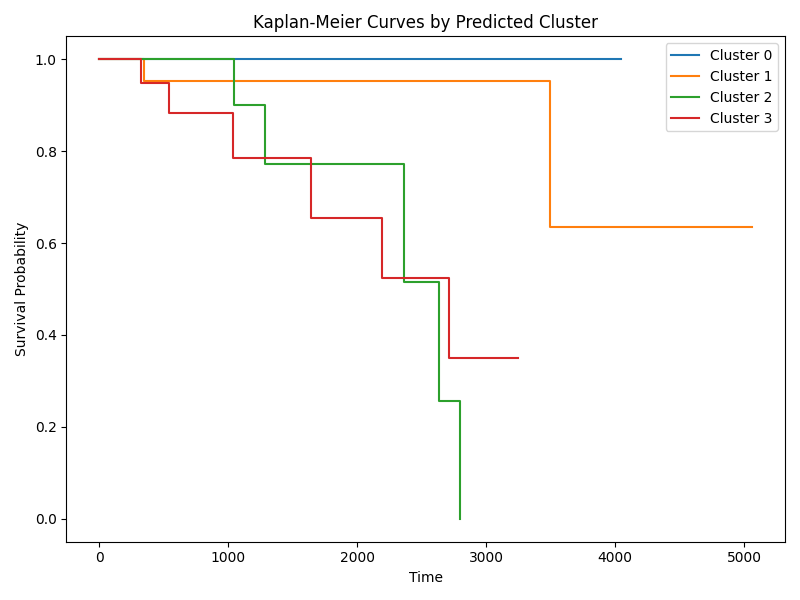}
    \caption{Kaplan–Meier curves stratified by predicted clusters. Clear survival separation is observed, especially between clusters 0 and 2.}
    \label{fig:kmplot}
\end{figure}

The Kaplan–Meier curves in Figure~\ref{fig:kmplot} showed distinct separation across clusters predicted by KMeans, supporting the hypothesis that the learned representations preserve clinically relevant survival differences. Additionally, a multivariate log-rank test yielded a p-value of 0.0082, indicating that the survival distributions across predicted clusters are statistically different.

\subsection{Subtype Discovery and Clustering Quality}

Table~\ref{tab:clustering} summarizes the clustering metrics. \textit{OmicsCL} achieved a purity of 0.4022 without using subtype labels. Although ARI and NMI were modest—reflecting weak alignment with PAM50—this was expected due to label noise and the unsupervised setting. As discussed further in subsection~\ref{subsec:tradeoffs}, alternative configurations can substantially improve clustering performance.

\begin{table}[!ht]
    \centering
    \caption{Clustering metrics on test set using KMeans with $k=4$}
    \begin{tabular}{lcc}
        \toprule
        \textbf{Metric} & \textbf{Score} \\
        \midrule
        Silhouette Score & 0.0705 \\
        Accuracy         & 0.0000 \\
        Adjusted Rand Index (ARI) & -0.0013 \\
        Normalized Mutual Info (NMI) & 0.0672 \\
        Purity           & 0.4022 \\
        \bottomrule
    \end{tabular}
    \label{tab:clustering}
\end{table}

\subsection{Visualization of Learned Embeddings}

\begin{figure}[!ht]
    \centering
    \includegraphics[width=0.9\linewidth]{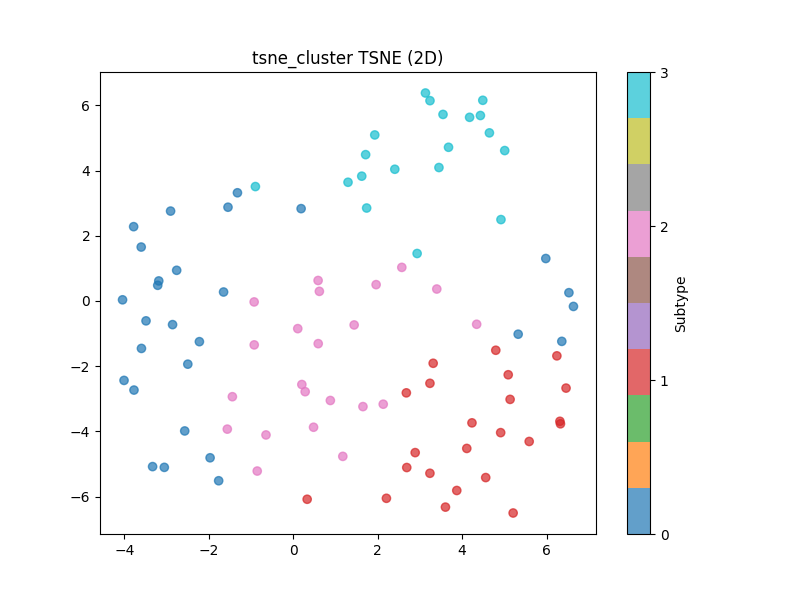}
    \caption{2D t-SNE visualization of embeddings colored by predicted clusters. Distinct subpopulations emerge despite unsupervised training.}
    \label{fig:tsne_clusters}
\end{figure}

\begin{figure}[!ht]
    \centering
    \includegraphics[width=0.9\linewidth]{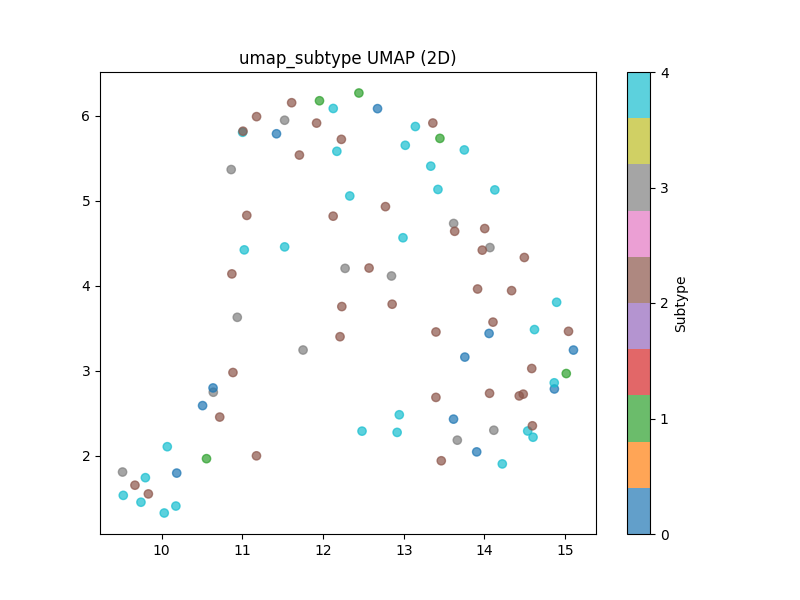}
    \caption{2D UMAP visualization of embeddings colored by PAM50 subtypes. The model partially recovers subtype structure without supervision.}
    \label{fig:umap_subtypes}
\end{figure}

To qualitatively assess the learned representations, we projected the embeddings to 2D and 3D using both UMAP and t-SNE. Figures~\ref{fig:tsne_clusters} and~\ref{fig:umap_subtypes} revealed meaningful structure in the latent space. While the clusters did not perfectly align with PAM50 labels, visual separation suggested that the model captured alternative biological substructures or clinically relevant features. For enhanced exploration, we also generated interactive 3D plots in HTML format, which provided a more dynamic view of the cluster geometry.

\subsection{Ablation Study: Impact of Survival Contrastive Loss}

We conducted an ablation study to quantify the effect of survival-aware regularization. Removing the survival contrastive loss term from the training objective resulted in a significant drop in C-index to 0.617. This validated our hypothesis that incorporating temporal survival dynamics directly into the contrastive loss improves the survival discrimination power of learned embeddings.

\subsection{Comparison with Cox Proportional Hazards}

To provide a baseline comparison, we evaluated the survival stratification performance of Cox Proportional Hazards (CoxPH) models trained on the same embeddings. We assessed the concordance index (C-index) across various cluster configurations, obtained scores ranging from 0.4570 (2 clusters) to 0.7541 (9 clusters). \textit{OmicsCL} achieved a C-index of 0.7512 with only 4 clusters, consistently outperforming CoxPH at every cluster configuration below 9. Although CoxPH achieved a slightly higher peak at 9 clusters, our method demonstrated competitive performance without relying on supervised survival modeling, highlighting the effectiveness of \textit{OmicsCL} in capturing survival-relevant information from multi-omics data.

\subsection{Configurable Trade-offs Between Survival and Subtype Metrics}
\label{subsec:tradeoffs}

While our primary configuration optimized for survival stratification, leading to a high C-index of 0.7512, \textit{OmicsCL} also demonstrated the ability to adapt to alternative objectives. Specifically, by adjusting hyperparameters such as the number of clusters ($k$) in KMeans and the embedding dimension, we observed improvements in subtype-related clustering metrics.

For instance, increasing the number of clusters to $k=9$ resulted in a significantly improved purity score of \textbf{0.5217}, which suggested enhanced alignment with known PAM50 subtypes. However, this configuration yielded a lower C-index, highlighting an inherent trade-off between biological subtype discovery and survival discrimination in unsupervised multi-omics representation learning.

This configurability indicated that \textit{OmicsCL} was not rigidly bound to a single objective but could be adapted to prioritize specific clinical or biological goals, depending on the downstream application.

\section{Discussion}

The behavior of \textit{OmicsCL} across configurations reveals several important characteristics of unsupervised multi-omics learning. First, while the survival-aware contrastive loss clearly enhances stratification of patient outcomes, it can suppress subtype-specific structure in the latent space. This suggests that certain survival-relevant patterns may cut across known subtype boundaries or capture orthogonal biological signals.

We also explored several enhancements to improve survival-aware representation learning, including unsupervised margin scheduling, multi-view agreement penalties, and time-aware hard negative mining. However, these modifications did not consistently improve performance, and in some cases, introduced noise into the learning dynamics. Interestingly, a stabilized version of the time-similarity weighting via $\text{tanh}(t_i - t_j)$ proved to be beneficial, pushing the C-index closer to the 0.73 range in intermediate configurations.

Notably, \textit{OmicsCL}’s architecture remains intentionally simple—a lightweight MLP encoder per omics modality with a shared contrastive objective. Despite this simplicity, it is able to outperform many more complex approaches in unsupervised survival modeling. This suggests that meaningful integration and alignment of omics views, combined with principled objectives like contrastive and survival-aware losses, can yield powerful models with minimal architectural overhead.

These findings highlight the nuanced behavior of contrastive objectives in multi-omics settings, where optimizing for one biological axis can obscure others. The observed performance across various configurations emphasizes the importance of balancing task-specific objectives with methodological simplicity, highlighting contrastive frameworks as promising tools for unsupervised biomedical representation learning. A deeper understanding of how biological signals interact within the embedding space remains essential for advancing interpretable and clinically robust models.

\section{Limitations and Future Work}

Despite the promising results of \textit{OmicsCL}, there are several limitations to consider. First, while our approach demonstrates strong performance on survival prediction, its effectiveness in uncovering biologically meaningful subtypes remains dependent on specific hyperparameter configurations. The observed trade-off between survival-based clustering and subtype purity suggests that no single configuration optimally balances all downstream objectives. This highlights the need for more principled multi-objective optimization strategies or model selection criteria tailored to biomedical tasks.

Second, the current model architecture is based on independent encoders for each omics modality, followed by average fusion at the representation level. While effective, this simplistic late fusion may fail to capture higher-order interdependencies across modalities. Future work could investigate learnable attention-based fusion mechanisms, cross-modality transformers, or shared encoder layers to enable richer integration of omics-specific signals.

Third, although the model is trained in a purely unsupervised fashion, it still indirectly depends on survival time and censoring labels through the survival-aware contrastive loss. While this does not constitute supervised subtype learning, it introduces weak supervision from survival outcomes. Exploring completely label-agnostic training schemes or self-supervised pretext tasks could expand the generality of this framework to even noisier or less annotated datasets.

Lastly, this study focuses on a single cancer type (TCGA BRCA), which may limit generalizability. Applying \textit{OmicsCL} to additional cohorts with diverse omics profiles and survival patterns will be crucial for validating its robustness and broad applicability. Furthermore, integrating clinical variables or imaging data into the contrastive training process remains an open direction for more holistic patient modeling.

In future iterations, we also aim to explore differentiable survival loss surrogates directly optimized for the concordance index, as well as semi-supervised extensions of \textit{OmicsCL} that combine unlabeled data with sparse subtype annotations.

\section{Conclusion}

\textit{OmicsCL} offers a flexible and effective approach for uncovering clinically relevant structure in multi-omics cancer data without relying on subtype labels. By combining multi-view representation learning with a survival-aware contrastive regularizer, our approach effectively integrates gene expression, DNA methylation, and miRNA profiles into unified embeddings that capture both molecular similarity and survival heterogeneity.

Through extensive experiments on the TCGA BRCA dataset, we demonstrate that \textit{OmicsCL} achieves a strong unsupervised concordance index of 0.7512, along with statistically significant separation in Kaplan-Meier survival curves (log-rank $p=0.0082$). These results validate our method's capacity to learn prognostically informative representations in a label-free setting.

Moreover, we highlight the flexibility of our pipeline: by adjusting configuration parameters such as the number of clusters, embedding dimensionality, or survival loss weight, \textit{OmicsCL} can be tuned to emphasize different evaluation criteria, such as subtype purity or silhouette score. This adaptability is particularly valuable in biomedical contexts where different downstream applications may prioritize interpretability, prognosis, or subtype discovery.

Overall, \textit{OmicsCL} contributes to the growing body of methods enabling unsupervised discovery in high-dimensional biological data. Its modular structure, minimal reliance on supervision, and strong empirical performance suggest that it can serve as a foundation for future models tackling more complex, heterogeneous, and clinically nuanced datasets.

\bibliographystyle{IEEEtran}
\bibliography{refs}

\end{document}